\documentclass[sigconf, screen]{acmart}
\AtBeginDocument{%
  }
    
\pdfoutput=1

\renewcommand\footnotetextcopyrightpermission[1]{}
\usepackage{algorithm}
\usepackage{algorithmic}
\usepackage{multirow}
\usepackage{bbding}
\usepackage{subcaption}

\begin{document}

\title{Emotion Knowledge Enhancement for Vision Large Language Models: A Self-Verification Approach for High-Quality Emotion Instruction Data Generation}

\author{Feifan Wang}
\affiliation{%
 \institution{Southeast University}
 \city{Nanjing}
 \country{China}}
 \email{aipl_wff@seu.edu.cn}

\author{Tengfei Song}
\affiliation{%
 \institution{Nanjing Research Center, Huawei Technologies Ltd}
 \country{China}}

 \author{Minggui He}
\affiliation{%
 \institution{Beijing Research Center, Huawei Technologies Ltd}
 \country{China}}
 
  \author{Chang Su}
\affiliation{%
 \institution{Beijing Research Center, Huawei Technologies Ltd}
 \country{China}}

  \author{Zhanglin Wu}
\affiliation{%
 \institution{Nanjing Research Center, Huawei Technologies Ltd}
 \city{Nanjing}
 \country{China}}

  \author{Hao Yang}
\affiliation{%
 \institution{Beijing Research Center, Huawei Technologies Ltd}
 \city{Beijing}
 \country{China}}

  \author{Wenming Zheng}
  \authornote{Corresponding author.}
\affiliation{%
 \institution{Southeast University}
 \city{Nanjing}
 \country{China}}

  \author{Osamu Yoshie}
\affiliation{%
 \institution{Waseda University}
 \city{Fukuoka}
 \country{Japan}}
 
\begin{abstract}
Facial emotion perception in the vision large language model (VLLM) is crucial for achieving natural human-machine interaction. However, creating high-quality annotations for both coarse- and fine-grained facial emotion analysis demands costly expertise. The lack of such high-quality instruction data limits the performance of VLLMs in facial emotion perception. To address this, we propose a self-verification approach with emotion knowledge enhancement (SEKE), which generates high-quality instruction data for multi-grained emotion analysis cost-effectively using closed-source VLLM. This approach integrates prior human knowledge to VLLM inference, guided by the inherent correlations between three grained levels of emotion descriptions, i.e., discrete expression, valence-arousal, and action unit, to reliably generate comprehensive annotations. A self-verification strategy with Uncertainty-Aware Monte Carlo sampling (SV-UAMC) is further embedded to efficiently extract more accurate VLLM predictions, further improving annotation reliability. Consequently, we construct a facial emotion instruction dataset (FEID) containing three comprehensive descriptions, which provides coarse- and fine-grained emotional information for effective model training. Additionally, we introduce a facial emotion analysis benchmark (FEAB) to measure the VLLM's corresponding ability. Our method significantly outperforms state-of-the-art methods on three downstream facial emotion analysis tasks.
\end{abstract}


\begin{teaserfigure}
\begin{center} 
  \includegraphics[width=6.6in]{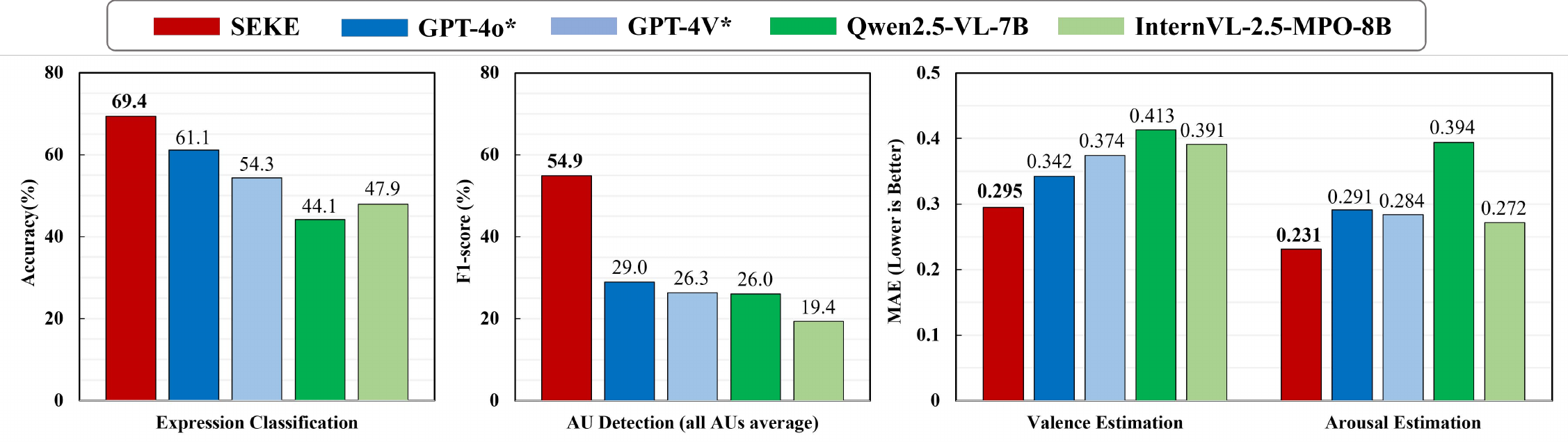}
\end{center}
  \caption{Benchmark performance of SEKE model.}
  \label{figteaser}
\end{teaserfigure}

\maketitle  

\section{Introduction}
\label{sec:intro}


Recent vision large language models (VLLMs) have shown substantial progress in human-robot interactions, with facial emotion analysis serving as a key foundation for generating natural responses \cite{li2024llava,openai2024gpt4o}. 
Human emotion is a sophisticated psychological state that encompasses physiological responses and behavioral expressions \cite{lindsley1951emotion}.
Specifically, human emotion can be described by different methods. 
One common method is the basic discrete emotion, which divides it into universal types such as happiness, anger, sadness, fear, surprise, and disgust \cite{ekman1987universals}. 
Another method is the valence-arousal (VA) model, which represents emotions on a continuous scale \cite{russell1980circumplex}. Valence reflects the positivity or negativity of an emotion, while arousal refers to the intensity of emotional activation. 
Additionally, for emotion analysis, facial action units (AUs) provide a more detailed description of facial muscle movements \cite{ekman1997face}. 
Given a specific emotional state, different representation methods have inherent correlations. For instance, a happy expression typically coincides with the activation of AU6 (Cheek Raiser) and AU12 (Lip Corner Puller), and the intensity is related to the arousal value. 
Each descriptive method captures distinct attributes of the same facial emotion \cite{liu2022facial}. 
However, each method alone offers an incomplete representation. Basic expressions fail to encode subtle intensity variations, while AUs lack emotional semantics without contextual cues. Similarly, VA dimensions ignore how emotions manifest through concrete facial muscles, e.g., high arousal may originate from AU5 (eye widening) in fear or AU25 (lips parting) in surprise. Therefore, integrating all three perspectives and maintaining their proper correlations is essential for analyzing the true emotion states, enabling cross-validation across coarse-to-fine granularities.

The outstanding performance of vision large language models (VLLMs) in various tasks can be largely attributed to the availability of vast amounts of high-quality instruction data \cite{openai2024gpt4o,kaplan2020scaling, Anthropic2024Claude}. This data provides models with a diverse and comprehensive understanding of complex tasks, enabling them to generalize well across different scenarios \cite{touvron2023llama}. 
However, manual annotation of instruction data for facial emotion analysis is both challenging and expensive. Emotions are inherently complex and often involve subtle expressions that require expert-level understanding to accurately label \cite{zhang2014bp4d}.
Moreover, large-scale datasets are needed to capture the full range of human emotions, which further increases the time and expense of manual annotation. 
Leveraging existing high-quality VLLMs to directly generate facial emotion instruction data offers a cost-effective strategy to alleviate reliance on manual annotations. However, while some VLLMs, such as GPT-4V \cite{openai2024gpt4V} and GPT-4o \cite{openai2024gpt4o}, demonstrate competence in recognizing basic emotions, they struggle with more fine-grained emotion analysis, including valence-arousal and AU. Lian et al. \cite{lian2024gpt} systematically assessed GPT-4V's performance across various emotion analysis tasks, substantiating the above conclusions with extensive experiments. 
In summary, fine-grained affective annotations such as VA and AUs demand specialized domain knowledge,  making manual labeling costly and VLLM labeling unreliable. This hinders the generation of large-scale instruction datasets that integrate all three facial emotion descriptions, limiting the fine-tuning of VLLMs for robust coarse and fine-grained facial emotion analysis.

This paper aims to tackle these challenges, achieving cost-effective generation of high-quality instruction data for VLLM facial emotion analysis, which is enriched with fine-grained emotional information.
We propose a self-verification approach with emotion knowledge enhancement (SEKE) to generate such data, which leverages human-annotated knowledge from open-source datasets and integrates this knowledge into the learning process of VLLMs.
Manual annotations on basic expression recognition, AU detection, and valence-arousal estimation provide high-quality knowledge for facial emotion analysis. 
However, most public datasets suffer from incomplete annotations.
Therefore, we employ VLLM to infer missing annotations based on the existing knowledge from manual annotations. 
This strategy integrates human expertise with the advanced reasoning capabilities of VLLMs, where human knowledge constrains the VLLM reasoning space to ensure generated annotations adhere to emotion manifestation mechanisms (e.g., AU12 activation rarely coexists with negative valence). In this way, a balance is achieved between human and VLLM annotations, maintaining efficiency while enhancing reliability.
Nonetheless, despite the knowledge guide, VLLM predictions may still exhibit hallucinations or unreliable results due to inherent uncertainties \cite{huang2024opera}.
Thus, a self-verification method with uncertainty-aware Monte Carlo sampling (SV-UAMC) is proposed to generate further reliable instructions in a low-cost manner, which estimates the epistemic uncertainties of VLLMs by Monte Carlo sampling. 
Epistemic uncertainty, stemming from limited knowledge or insufficient statistical evidence, objectively reflects model performance \cite{kendall2017uncertainties}. 
By dynamically prioritizing predictions with low epistemic uncertainty during Monte Carlo sampling, which indicates higher VLLM reliability in limited-knowledge scenarios, SV-UAMC efficiently filters the most reliable annotations from minimal repetitions.
With the generated reliable annotations, VLLMs can provide more accurate descriptions for facial emotion analysis, which stems from the reason that comprehensive annotations enable VLLMs to effectively integrate coarse- and fine-grained information while capturing their underlying correlations.
Finally, we release a facial emotion instruction dataset (FEID) and a facial emotion analysis benchmark (FEAB). 
As we know, this is the first high-quality instruction dataset with comprehensive coarse- and fine-grained emotional descriptions alongside their correlation reasoning knowledge for VLLM emotion analysis.

Our contributions can be summarized as follows:

\begin{itemize}
\item We propose a \textbf{SEKE} method to generate high-quality facial emotion instruction data, which integrates human prior knowledge and uncertainty-aware self-verification strategy to improve the annotation reliability of large models.
\item We construct an instruction dataset \textbf{FEID} and a benchmark \textbf{FEAB}, where FEID provides each sample with all three descriptions and their correlations, significantly enhancing VLLM’s ability to perceive fine-grained emotional cues.
\item The model fine-tuned on our FEID significantly outperforms state-of-the-art models on three downstream facial emotion analysis tasks, as shown in Figure \ref{figteaser}.
\end{itemize}
\section{Related Works}
\label{sec:related}

\subsection{Vision Large Language Model}
\label{sec:Related Works:Emotion large language model}

Under the guidance of the scaling laws \cite{kaplan2020scaling}, vision LLMs, such as GPT-4V \cite{openai2024gpt4V}, GPT-4o \cite{openai2024gpt4o}, and Claude-3.5 \cite{Anthropic2024Claude}, integrate vision and language understanding and demonstrate outstanding performance across various vision tasks. 
Li et al. \cite{li2024llava2} introduced an open-source LLAVA-OneVision, developed from the Large Vision-and-Language Assistant (LLaVA) \cite{liu2024visual}, which is capable of effectively handling image and video scenarios. 
For facial emotion analysis, Lian et al. \cite{lian2024affectgpt} leveraged the GPT-4V along with several multimodal large language models to generate emotion instructions and developed AffectGPT \cite{lin2023video}. However, due to the lack of high-quality expert emotion annotations, AffectGPT is limited to performing basic emotion recognition and description tasks. 
Furthermore, Cheng et al. \cite{cheng2024emotion} utilized OpenFace to estimate discrete emotion labels and AU intensities, which were then used to generate corresponding contextual descriptions. Training on these generated data, Emotion-LLAMA \cite{cheng2024emotion} is proposed to integrate audio, visual, and textual inputs for emotion analysis. All these MLLMs have enhanced emotion recognition and reasoning capabilities. However, these methods offer only a rough description, and more fine-grained facial emotion analysis, such as AU and valence-arousal descriptions, are needed. To address this, we propose a method to generate high-quality instruction data for more fine-grained facial emotion analysis.

\begin{figure*}[tb]
\begin{center} 
\includegraphics[width=6.5in]{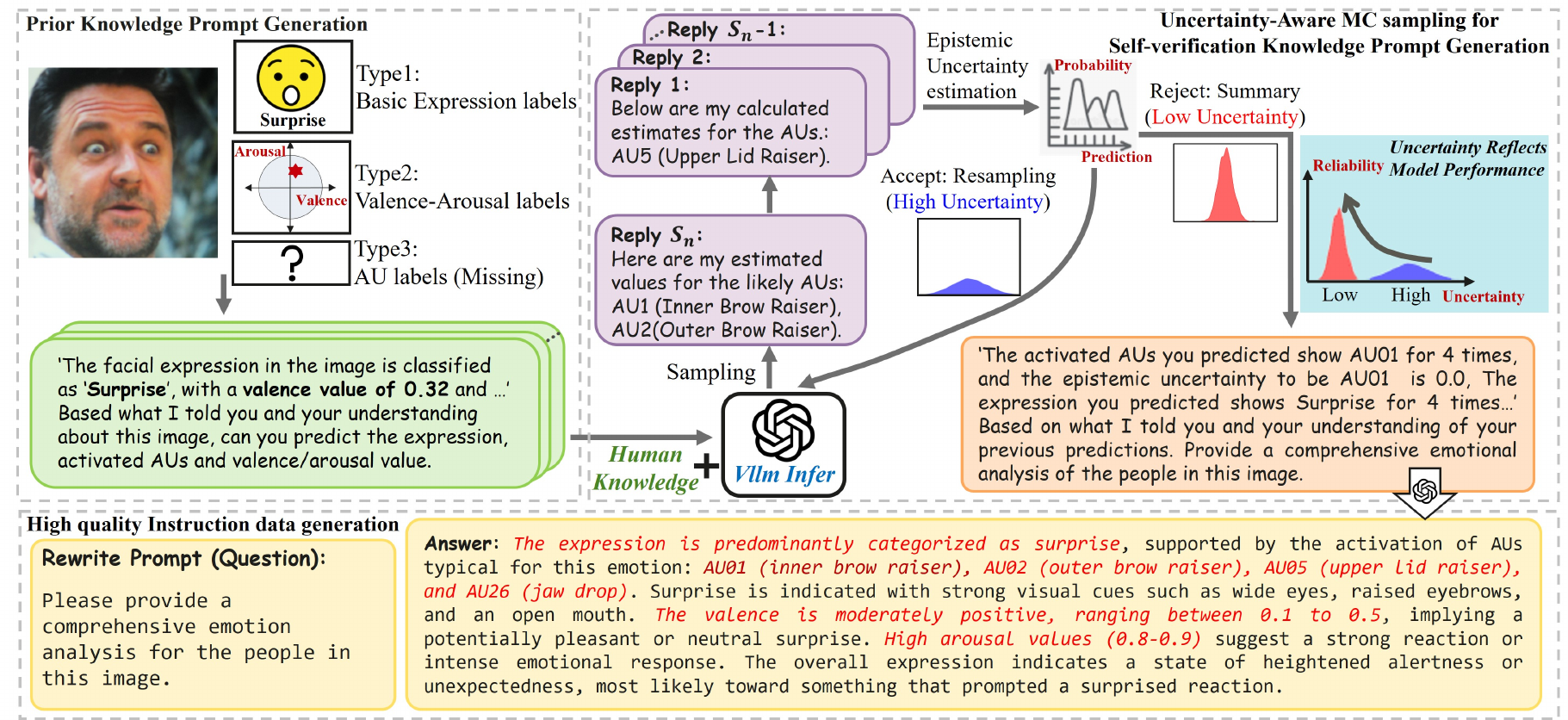}
\end{center}
\caption{The overall architecture of the proposed self-verification approach with emotion
knowledge enhancement (SEKE) to generate high-quality instruction data for facial emotion analysis.}
\label{all}
\vspace{-0.3cm}
\end{figure*}

\subsection{Facial Emotion Analysis}
\label{sec: Related Works: Facial Emotion Analysis}

Facial expression is one of the most crucial modalities for humans to convey emotions, and a growing number of researchers are focusing on improving the accuracy of tasks related to facial emotion analysis \cite{li2020deep, Song_2021_CVPR, NEURIPS2020_a51fb975,adyapady2023comprehensive, Yang_2024_CVPR}. 
Cui et al. \cite{cui2020knowledge} embed emotion knowledge for joint expression and action unit recognition in an unsupervised manner. 
Paskaleva et al. \cite{paskaleva2024unified} proposed a unified and interpretable emotion model for expression generation that simultaneously accounts for discrete emotion categories, AU labels, and valence-arousal values. This unified expression description method provides a consistent representation. Most previous works focus on one type of facial emotion description method, but different descriptions provide complementary information. In this paper, we integrate different types of description knowledge to achieve comprehensive facial emotion analysis, capturing both coarse- and fine-grained emotional cues.

\subsection{Facial Emotion Analysis Data Generation}

Many research studies have contributed significantly to the field of facial emotion perception by releasing numerous datasets specifically designed for downstream tasks.  
Zhang et al. \cite{zhang2014bp4d} introduced the Binghamton-Pittsburgh 4D Facial Expression Database (BP4D). A notable strength of the BP4D dataset is its comprehensive AU annotations. 
Aff-wild2 \cite{kollias2019expression} includes manual annotations for expression, AUs, and valence/arousal values,  providing comprehensive emotion descriptions. 
Although these datasets contain precise human-annotated ground truth labels, their scale is far from sufficient to harness the potential of VLLMs, particularly the fine-grained emotion annotations such as AUs and valence-arousal values.
Thus, more studies are concentrating on automatically generating large-scale emotion instruction datasets. 
Emotion-LLama leverages open-source large models like MiniGPT-v2 \cite{chen2023minigpt} to annotate numerous unlabeled data. A subset of this data is further analyzed using LLaMA-3 \cite{llama3modelcard} to develop the MERR-Fine dataset \cite{cheng2024emotion}, which encompasses comprehensive reasoning knowledge for integrating multimodal cues. However, these large models are not specifically designed for emotion tasks, and the reliability is not validated or improved. 
Xie et al. \cite{xie2024emovit} leverage GPT-4 to automatically generate diverse queries based on manually annotated emotional attributes, constructing the Emovit. Its diverse instructions have enhanced the model's performance. However, Emovit is not human-centered, omits fine-grained human emotion descriptions, and depends entirely on manual annotations. 
To develop emotion instruction datasets with comprehensive coarse- and fine-grained cues, we propose a strategy that integrates human prior knowledge with VLLM self-verification, enabling efficient and reliable generation of missing annotations beyond manual labeling.

\section{Methods}

\subsection{The overall architecture of data generation}
As shown in Figure \ref{all}, the overall architecture of the data generation process consists of three parts, i.e., prior knowledge prompt generation, uncertainty-aware MC sampling for self-verification knowledge prompt generation, and high-quality instruction data generation. 

For the first part, we leverage the manually annotated emotion labels from open-source datasets as prior knowledge. Since the annotations are generally incomplete, we design prompts to embed this prior knowledge and predict the missing descriptions. By incorporating meticulously curated human knowledge with the inferring capabilities of VLLMs, which are guided by the inherent correlations between emotion descriptions, we can reliably fill in the annotation gaps. As an example shown in Figure \ref{all}, we leverage the annotated discrete basic expressions and valence-arousal values to predict activated AUs.

Further, to obtain more reliable results, we estimate the epistemic uncertainties of VLLM output by generating multiple results. Then, we analyze the results from multiple samples and their corresponding uncertainties as knowledge, embedding them into prompts to generate more reliable outcomes. Since epistemic uncertainty is related to limited knowledge and insufficient statistical evidence, it provides an effective way for VLLM to estimate which parts are reliable and which are not, enabling the generation of more dependable results. To minimize costs, uncertainty-aware MC sampling is introduced to adaptively adjust the number of samples in this part.

Finally, VLLM outputs a high-quality emotion analysis without additional manual annotations. 
This analysis not only captures three comprehensive emotion descriptions spanning coarse- to fine-grained cues but also contains inferring of correlations between them, enhancing the VLLM for accurate affective reasoning. 
A rewrite prompt is generated as the final question, devoid of any answer information. The prompt is randomly selected from the 11 templates we designed to enhance diversity, as detailed in Appendix A. In this paper, we leverage GPT-4o \cite{openai2024gpt4o} as the VLLM to generate instruction data.

\subsection{Uncertainty-Aware Self-Verification Data Generation with Embedded Knowledge}

As illustrated in Figure \ref{all}, we proposed a self-verification method to integrate prior knowledge and predict the unknown emotion labels. Specifically, we employ a probabilistic method by estimating the epistemic uncertainties to obtain more reliable results.

Probability uncertainty \cite{depeweg2018decomposition} comprises two types: aleatoric uncertainty and epistemic uncertainty. Aleatoric uncertainty captures the inherent data noise, while epistemic uncertainty refers to the uncertainty within the model, arising from limited knowledge or insufficient statistical evidence. The epistemic uncertainties are closely linked to the model's performance, where high-uncertainty regions often correlate with unreliable outputs due to extrapolation beyond training distributions or ambiguous context.
Thus, the purpose of using a self-verification approach based on epistemic uncertainty is to enable the model to distinguish between reliable and unreliable parts. This allows the model to further generate more dependable emotion descriptions.

To provide a clear representation, we use one-turn representation, the prediction of LLM can be represented by
\begin{equation}
\label{eq8}
\begin{aligned}
p(y|\bold{I}_{v}, \mathbf{I}_{t})&=\mathbb{E}_{\mathcal{E}\sim p(\mathcal{E}|\mathcal{D})}[p(y|\bold{I}_{v}, \mathbf{I}_{t}, \mathcal{E})]\\
&=\int p(y_{i}|\bold{I}_{v}, \mathbf{I}_{t}, \mathcal{E})p(\mathcal{E}| \bold{I}_{v}, \mathbf{I}_{t}) \ d\mathcal{E}\\ &\approx\frac{1}{S}\sum_{s=1}^{S}f( \bold{I}_{v},\mathbf{I}_{t},\mathcal{E}), 
\end{aligned}
\end{equation}
in which $y$ denotes the output of VLLM for specific prediction term, $\mathcal{E}$ represents the random variables, and $S$ is the number of samplings.

\begin{algorithm}[t]
\caption{Uncertainty-Aware Monte Carlo sampling}
\label{algorithm1}
\begin{algorithmic}[1]
\REQUIRE 
Facial images $I_v$ with partial annotations \\
\ENSURE 
Prompt and corresponding reliable answers about expression, valence-arousal values, activated AUs, and a comprehensive description \\ 
\STATE Initialize model parameters, generate prior knowledge prompt\\
\STATE Sampling 2 times, with one response inferred each time, $S_n=2$ \\
\WHILE{$S_n< N$}
        \STATE Estimate the uncertainty $\mathcal{U}(y_{n,i})$ \\
        \STATE Normalize the uncertainty $\Bar{\mathcal{U}}(y_{n,i})$ \\
        \STATE Select the maximize uncertainty $\Bar{\mathcal{U}}(y_{n,max})$ \\
        \STATE Estimate accept probability: $p_{acc}$\\
        \STATE Generate random value $p$ in [0,1]
            \IF{$p < p_{acc}$}
                \STATE Sample one more time, $S_n=S_n+1$
            \ELSE
                \STATE $break$
            \ENDIF
\ENDWHILE
\STATE Summary self-verification knowledge prompt by VLLM\\
\RETURN Generated instruction data
\end{algorithmic}
\end{algorithm}

For specific prediction tasks, the epistemic uncertainty can be approximated as
\begin{equation}\label{probability-acc}
\begin{aligned}
\mathcal{U}(y_i) &\approx  \mathbb{E}_{p(\mathcal{E}|\mathcal{D})}[f( \bold{I}_{v},\mathbf{I}_{t},\mathcal{E})^2] -(\mathbb{E}_{p(\mathcal{E}|\mathcal{D})}[f( \bold{I}_{v},\mathbf{I}_{t},\mathcal{E})])^2
\\
&=\frac{1}{S}\sum_{s=1}^{S} f( \bold{I}_{v},\mathbf{I}_{t},\mathcal{E}^{s})^{2} - (\frac{1}{S}\sum_{s=1}^{S}f( \bold{I}_{v},\mathbf{I}_{t},\mathcal{E}^{s}))^2 ,
\end{aligned}
\end{equation}
in which $y_i$ denote the prediction of the $i$-th task among the three emotion descriptions. The prediction should be sampled  $S$  times, after which the uncertainties for different tasks are summarized.

Estimating epistemic uncertainty typically requires extensive sampling, which can be costly. To accurately characterize unreliable results while reducing sampling costs, we focus on resampling high-uncertainty samples and minimizing resampling for low-uncertainty ones. Inspired by this, we propose Uncertainty-Aware Monte Carlo (UAMC) sampling, a method that estimates epistemic uncertainties in a cost-efficient manner.

Instead of using a fixed sampling number $S$, we implement a dynamic sampling number $S_n$ for each particular sample $n$. As demonstrated in Algorithm \ref{algorithm1}, our initial approach involves sampling twice with the prior knowledge prompt. For subsequent sampling steps, we estimate the uncertainties with the predicted labels. 
The epistemic uncertainty with UAMC sampling can be estimated by 
\begin{equation}\label{probability-acc}
\begin{aligned}
\mathcal{U}(y_{n,i}) &\approx  \mathbb{E}_{p(\mathcal{E}|\mathcal{D})}[f( \bold{I}_{v},\mathbf{I}_{t},\mathcal{E})^2] -(\mathbb{E}_{p(\mathcal{E}|\mathcal{D})}[f( \bold{I}_{v},\mathbf{I}_{t},\mathcal{E})])^2
\\
&=\frac{1}{S_n}\sum_{s=1}^{Sn} f( \bold{I}_{v},\mathbf{I}_{t},\mathcal{E}^{s})^{2} - (\frac{1}{S_n}\sum_{s=1}^{S_n}f( \bold{I}_{v},\mathbf{I}_{t},\mathcal{E}^{s}))^2 ,
\end{aligned}
\end{equation}
where $S_n$ represents the current cumulative number of samplings. Subsequently, the uncertainty $\mathcal{U}(y_{n,i})$ is normalized to fall within the range of 0 to 1, denoted as $\Bar{\mathcal{U}}(y_{n,i})$. Here, $i$ refers to the index of different tasks. We then select the highest uncertainty value, denoted $\Bar{\mathcal{U}}(y_{n,max})$, among different tasks. Further, the accept probability is calculated by 
\begin{equation}\label{probability-acc2}
\begin{aligned}
p_{acc}= \frac{1}{2}+\frac{1}{2}\Bar{\mathcal{U}}(y_{n,max}).
\end{aligned}
\end{equation}
Instead of setting a threshold, we utilize a robust approach by drawing from an acceptance probability $p_{acc}$ to decide on continuing sampling. If the prediction exhibits high uncertainty, it will have a high probability of acceptance for continued sampling, whereas low uncertainty likely leads to termination. This strategy dynamically allocates resources to focus on ambiguous cases and minimize redundant iterations for confident predictions, thereby improving efficiency. $N$ represents the maximum number of samplings.

Upon completing the sampling process, the prior knowledge, predicted results, and uncertainty statistics are combined to form a new summary prompt. This prompt is then fed into VLLM to produce high-quality emotion instruction data, which we call Facial Emotion Instruction Dataset (FEID). 
Our FEID incorporates three comprehensive emotional descriptions, which provide both coarse- and fine-grained clues. Additionally, the joint affective reasoning across these descriptions is also included, enabling the fine-tuned model to enhance facial emotion analysis by triangulating multi-level evidence.

\begin{figure}[t]
\begin{center} 
\includegraphics[width=3in]{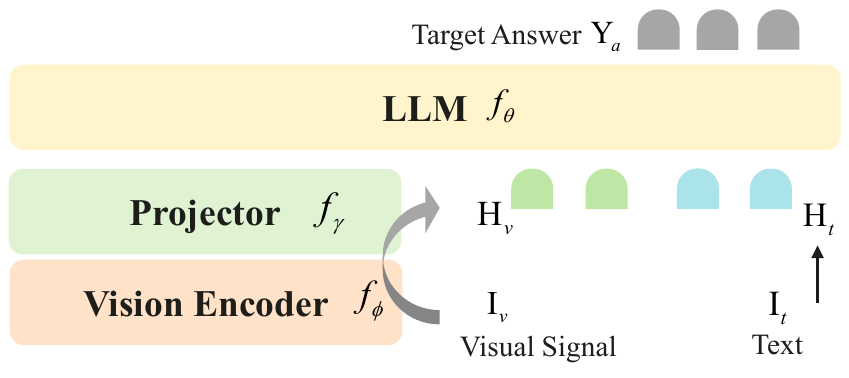}
\end{center}
\caption{The VLLM architecture used to train our model on the proposed facial emotion instruction dataset.}
\label{vllm}
\vspace{-0.3cm}
\end{figure}

\subsection{Emotion Instruction Tuning of VLLM}
In this paper, we use LLAVA-OneVision \cite{li2024llava2} as our baseline model to perform supervised fine-tuning (SFT) with our generated emotion instruction dataset.
The network architecture of our VLLM in this paper is illustrated in Figure \ref{vllm}. The model consists of a vision encoder $f_{\phi}$, a projector $f_{\gamma}$ and a language model $f_{\theta}$. The vision encoder $f_{\phi}$ and large language model $f_{\theta}$ are from siglip \cite{zhai2023sigmoid} and Qwen-2 \cite{yang2024qwen2}. The projector $f_{\gamma}$ is a 2-layer MLP \cite{liu2024improved}.

Given an input image $\bold{I}_{v}$, visual features can be computed as $\bold{F}_v=f_{\phi}(\bold{I}_{v})$. Then, a 2-layer MLP projector projects visual features to visual tokens $\bold{H}_v = f_{\gamma}(\bold{F}_v)$. The input text $\bold{I}_{t}$ is processed through a tokenizer $\bold{T}$, yielding the text tokens $\bold{H}_{t}=\bold{T}(\bold{I}_{t})$. Both visual tokens and text tokens are subsequently fed into LLMs. For an $L$-turn conversation, the target answers $\bold{Y}_{a}$ can be estimated as 
\begin{equation}\label{probability-acc}
 p(\bold{Y}_{a}|\bold{I}_{v},\bold{I}_{t}) = \prod_{l=1}^{L} p(y_l | \bold{I}_{v}, \mathbf{I}_{t, <l}, \mathbf{I}_{a, <l}),
\end{equation}
in which $\mathbf{I}_{t, <l}$ and $\mathbf{I}_{a, <l}$ represent the instructions and answers from all previous turns before the current prediction token $y_l$.  To handle higher-resolution images, the AnyRes \cite{li2024llava2} scheme is employed. Then, the objective is to maximize the likelihood of generating target answers $p(\bold{Y}_{a}|\bold{I}_{v},\bold{I}_{t})$. The LLaVA-OneVision fine-tuned on our SEKE-generated FEID is referred to as the SEKE model in subsequent sections.

\section{Experiment}

\setlength{\abovecaptionskip}{2pt}

\begin{table}
\centering
\renewcommand{\arraystretch}{1.1}
\tabcolsep0.13in
\caption{Statistics of FEID and FEAB. EXP represents the discrete expression label, VA represents the Valence/Arousal label, and AU represents the facial AU detection label.‘Train’ and ‘Test’ indicate the number of images used to construct FEID and FEAB.}
\begin{tabular}{c|cc|c}
\hline
\hline
\multirow{2}{*}{\textbf{Dataset}} & \multicolumn{2}{c|}{\textbf{Samples}} & \multirow{2}{*}{\textbf{Label}} \\ \cline{2-3}
 & \multicolumn{1}{c|}{\textbf{Train}} & \textbf{Test} &  \\ \cline{2-3} \hline
CK+  \cite{lucey2010extended} & \multicolumn{1}{c|}{293} & 34 & EXP \\ \hline
RAF-DB \cite{li2017reliable} & \multicolumn{1}{c|}{2453} & 608 & EXP \\ \hline
AffectNet \cite{mollahosseini2017affectnet} & \multicolumn{1}{c|}{5752} & 496 & EXP, VA \\ \hline
DISFA \cite{mavadati2013disfa} & \multicolumn{1}{c|}{7001} & 716 & AU \\ \hline
BP4D \cite{zhang2013high} & \multicolumn{1}{c|}{6260} & 689 & AU \\ \hline
Aff-Wild2 \cite{kollias2019expression} & \multicolumn{1}{c|}{4479} & 492 & EXP, VA, AU \\ \hline
\textbf{FEID}  & \multicolumn{1}{c|}{26238} & - & EXP, VA, AU \\ \hline
\textbf{FEAB}   & \multicolumn{1}{c|}{-} & 3035 & EXP, VA, AU \\ \hline
 \hline
\end{tabular}
\vspace{-0.3cm}
\label{table1}
\end{table}

\subsection{The FEID and the FEAB}

This section introduces our generated instruction dataset FEID and Facial Emotion Analysis Benchmark (FEAB). They are proposed to train a high-quality facial emotion VLLM and comprehensively test its emotion perception capabilities, respectively. The datasets achieve full coverage of three key emotion descriptions, i.e., discrete emotion, valence-arousal, and facial AU. We construct train and test data by sourcing independent subject samples from six popular emotion analysis datasets, i.e., CK+ \cite{lucey2010extended}, RAF-DB \cite{li2017reliable}, AffectNet \cite{mollahosseini2017affectnet}, DISFA \cite{mavadati2013disfa}, BP4D \cite{zhang2013high}, and Aff-Wild2 \cite{kollias2019expression}. Table \ref{table1} presents the data statistics for FEID and FEAB, including the number of images from other datasets. More details of our data are categorized by types of emotional descriptions as follows:

\noindent\textbf{Discrete Emotion Recognition:} We select a portion of data from CK+, RAF-DB, AffectNet and  Aff-Wild2 as raw data for expression recognition. For CK+, following prior studies \cite{jiang2022disentangling}, we use the last frame of each video with the peak expression intensity. 
RAF-DB, AffectNet, and Aff-Wild2 are all in-the-wild image datasets sourced from the internet. We randomly extract partial samples from each entire database, simultaneously ensuring a balanced representation across eight categories.

\noindent\textbf{Valence-Arousal Estimation } We select AffectNet and Aff-Wild2 as data sources in valence-arousal estimation. These samples are all annotated with valence and arousal values ranging from [-1, 1].

\noindent\textbf{Facial AU Detection } We gather data for AU detection from BP4D, DISFA, and Aff-Wild2, each of which includes 12 AU categories. In total, we have 17 AU categories, as shown in Table \ref{table2}.
DISFA also provides AU intensity annotations, where AUs with intensities higher than 2 are considered occurrences.
Based on these raw data, we further separate them into FEID and FEAB.

For FEAB, we select 3035 images from the raw data in an independent subject manner to avoid information overlap with the train set. This is achieved by sampling 1/10 of the subjects from each raw dataset. FEAB's labels are derived solely from manual annotations, ensuring fair evaluation.

For FEID, we have 26238 remaining images, which are then annotated through our SEKE method to generate comprehensive instructions. Each sample is labeled with all three emotional descriptions, partly coming from the manual annotations and partly supplemented using SEKE. The sample distribution of each emotional description in the manually annotated FEID part is presented in Appendix B. 

\begin{table*}[t]
\small
\centering
\renewcommand{\arraystretch}{1.4}
\tabcolsep0.02in
\caption{Evaluation results of our SEKE and the state-of-the-art methods on FEAB. The best result for each description is highlighted in bold. The larger the accuracy value for expression and the F1 score for AU, the better. The smaller the MAE values for valence and arousal, the better. * denotes closed-source models, and others are open-source models.}
\begin{tabular}{l|c|ccccccccccccccc|c|c}
\hline
\hline
\multirow{2}{*}{Method} & EXP & AU1 & AU2 & AU4 & AU6 & AU7 & AU10 & AU12 & AU14 & AU15 & AU17 & AU23 & AU24 & AU25 & AU26 & All & Valence & Arousal \\ \cline{2-19} 
 & Acc (\%) & \multicolumn{15}{c|}{F1 score (\%)} & MAE & MAE \\ \hline
LLaVA-NeXT-Interleave \cite{li2024llava33} & 36.8 & 26.0 & 18.2 & 14.5 & 18.3 & 3.1 & 7.6 & 0.2 & 0.0 & 0.0 & 0.0 & 0.0 & 0.0 & 0.0 & 0.0 & 6.3 & 0.423 & 0.310 \\
MiniCPM-V2.6 \cite{yao2024minicpm} & 30.2 & 10.6 & 4.1 & 46.9 & 53.3 & 28.3 & 45.7 & 57.8 & 9.3 & 3.9 & 2.4 & 2.4 & 1.5 & 10.7 & 13.5 & 20.7 & 0.472 & 0.351 \\
Emotion-LLaMA \cite{cheng2024emotion} & 41.8 & 26.3 & 19.2 & 43.3 & 51.2 & 48.7 & 43.0 & 57.4 & 31.7 & \textbf{16.6} & 32.4 & 14.9 & 9.0 & 32.5 & 4.5 & 30.8 & 0.476 & 0.318 \\
LLaVA-OneVision \cite{li2024llava2} & 47.6 & 22.1 & 21.1 & 55.8 & 55.8 & 45.3 & 43.5 & 49.3 & 33.9 & 15.1 & 30.0 & 16.9 & 9.2 & 41.1 & 12.9 & 32.3 & 0.443 & 0.327 \\
InternVL-2.5-MPO-8B \cite{wang2024enhancing} & 47.9 & 19.7 & 16.2 & 7.6 & 61.8 & 0.0 & 52.5 & 67.3 & 2.0 & 12.3 & 18.3 & 3.2 & 0.0 & 1.1 & 9.8 & 19.4 & 0.391 & 0.272 \\
Qwen2.5-VL-7B-Instruct \cite{bai2025qwen2} & 44.1 & 22.9 & 12.7 & 52.4 & 42.4 & 20.1 & 40.4 & 51.2 & 23.3 & 9.1 & 25.7 & 6.7 & 5.4 & 37.4 & 14.2 & 26.0 & 0.413 & 0.394 \\
GPT-4V* \cite{openai2024gpt4V} & 54.3 & 14.9 & 12.8 & 69.9 & \textbf{69.3} & 8.2 & 11.4 & 78.3 & 2.2 & 9.7 & 14.7 & 2.2 & 0.0 & 41 & 33 & 26.3 & 0.374 & 0.284 \\
GPT-4o* \cite{openai2024gpt4o} & 61.1 & 33.8 & 32.1 & 69.0 & 65.6 & 9.3 & 10.8 & 73.7 & 4.4 & 13.0 & 27.4 & 2.6 & 17.2 & 26.3 & 20.5 & 29.0 & 0.342 & 0.291 \\ \hline
\textbf{SEKE} & \textbf{69.4} & \textbf{55.1} & \textbf{44.5} & \textbf{72.3} & 65.6 & \textbf{69.1} & \textbf{60.3} & \textbf{78.6} & \textbf{65.7} & 13.4 & \textbf{47.8} & \textbf{36.4} & \textbf{46.8} & \textbf{83.4} & \textbf{29.5} & \textbf{54.9} & \textbf{0.295} & \textbf{0.231} \\
\hline
\hline
\end{tabular}
\label{table2}
\vspace{-0.3cm}
\end{table*}

\subsection{Experimental Setup}

\noindent\textbf{Implementation Details}
During data generation, we set $N=5$ as the maximum sampling number to estimate uncertainties. During training, we use LLaVA-OneVision\cite{li2024llava2} as our base model to conduct full parameter SFT using our proposed FEID. The learning rate is set to 1e-5.
All experiments are implemented with PyTorch.

\noindent\textbf{Evaluation Metrics} 
Accuracy is used to evaluate the performance of basic facial expression classification.
The F1 score is provided to evaluate the performance of AU detection, particularly suited for binary classification tasks with imbalanced data.  Given the precision $p$ and the recall $r$, F1 score is calculated by $\text{F1}=2\frac{p \cdot r}{p+r}$. We report the F1 score specifically for the positive class. We employ the mean absolute error (MAE) for the evaluation of the valence/arousal estimation. It quantifies the average magnitude of the error by taking the square root of the average squared differences between predicted and actual values. To assess the model performance after SFT, we used an LLM to process its output. By fixing the outputs from different models into a universal template, it is easier to extract the numerical results of three tasks for metics computing. See Appendix C for more details.

\subsection{Compared with the State-of-the-art Methods}
To evaluate the performance of our proposed methods, we utilize two closed-source models, i.e., GPT-4V \cite{openai2024gpt4V} and GPT-4o \cite{openai2024gpt4o}, and six open-source models, i.e., LLaVA-NeXT-Interleave\cite{li2024llava33}, MiniCPM-V2.6 \cite{yao2024minicpm}, Emotion-LLaMA \cite{cheng2024emotion}, LLaVA-OneVision \cite{li2024llava2}, InternVL-2.5-MPO-8B \cite{wang2024enhancing}, and Qwen2.5-VL-7B-Instruct \cite{bai2025qwen2} for facial emotion analysis on FEAB,  with the results presented in Table \ref{table2}. 

The proposed SEKE model achieves the best performance across three facial emotion analysis tasks. Although GPT-4o was used as the VLLM to generate facial emotion instruction data, our self-verification process, enhanced with prior facial emotion knowledge, yielded significantly improved results. Specifically, SEKE outperforms GPT-4o with an 8.3$\%$ improvement in expression recognition, a 25.9$\%$ increase in AU detection accuracy, and reductions of 0.047 and 0.06 in MAE for valence and arousal estimation, respectively. Emotion-LLaMA is a model designed for video emotion analysis, fine-tuned on the emotion instruction dataset MER-Fine. However, it lacks emphasis on finer-grained emotional descriptions and their correlations, exhibiting limitations in discerning emotions across coarse- and fine-grained levels. 
While InternVL-2.5 and Qwen2.5-VL represent state-of-the-art VLLMs trained on large-scale instruction data, the lack of emotion-specific knowledge in the data limits their proficiency in emotion tasks compared to our SEKE model.
The marked performance of the SEKE model demonstrates that our generated instruction data,  the first to incorporate comprehensive and sufficient emotional descriptions, provides a viable strategy for advancing VLLM emotion perception capabilities.

\begin{table}[tb]
\centering
\caption{Ablation study of different instruction datasets. Expression accuracy ($\%$), average F1 score for AU detection ($\%$), and MAE for valence/arousal estimation on FEAB.}
\renewcommand{\arraystretch}{1.4}
\tabcolsep0.025in
\begin{tabular}{l|c|c|c|c}
\hline
\hline
\multirow{2}{*}{Dataset} & EXP & AU All & Valence & Arousal \\ \cline{2-5} 
 & Acc & F1 score & MAE & MAE \\ \hline
EmoVIT & 41.2 & 28.6 & 0.420 & 0.541 \\
MAFW & 52.9 & 24.4 & N/A & N/A \\
MERR-Fine & 37.1 & 24.7 & N/A & N/A \\ \hline
\textbf{FEID} & \textbf{69.4} & \textbf{54.9} & \textbf{0.295} & \textbf{0.231} \\
\hline
\hline
\end{tabular}
\vspace{-0.5cm}
\label{table3}
\end{table}

\subsection{Ablation Study} 

\subsubsection{Ablation Study of Different Instruction Dataset}
To evaluate the efficacy of our FEID, Table \ref{table3} compares the performance of models fine-tuned with different instruction datasets. For consistency, LLaVA-OneVision is standardized as the same base model, which also supports video data fine-tuning in MERR-Fine. 

The notable performance of FEID underscores the limitations of existing emotion instruction datasets, i.e., they inadequately address fine-grained visual emotional cues like AUs. Models fine-tuned on MAFW \cite{liu2022mafw} and MERR-Fine struggle to generate reliable valence-arousal estimations. 
While Emovit generates predictions, its lack of human-centric design limits its effectiveness in analyzing such fine-grained human emotions.
Although MAFW manually annotates descriptive labels, only a limited portion pertains to emotion reasoning, as this requires specialized expertise. MERR-Fine primarily relies on automated annotation via general-purpose large models, and the reliability is not guaranteed since they are not emotion-specific.
In contrast, our SEKE generation method strikes a balance between manual and large-model labeling. Integrating the knowledge guidance from prior manual annotations and the uncertainty-aware self-verification strategy, it enhances the reliability of large-model annotations without introducing additional labor. This makes FEID an instruction dataset with sufficient and reliable fine-grained emotion descriptions, fully enhancing VLLM emotion perceptions.

\begin{table}[tb]
\centering
\caption{Ablation study of different data generation methods. Expression accuracy ($\%$), average F1 score for AU detection ($\%$), and MAE for valence/arousal estimation on FEAB.}
\renewcommand{\arraystretch}{1.4}
\tabcolsep0.025in
\begin{tabular}{cc|c|c|c|c}
\hline
\hline
\multicolumn{2}{c|}{Method} & EXP & AU All & Valence & Arousal \\ \hline
Prior & \multicolumn{1}{l|}{SV-UAMC} & Acc & F1 score & MAE & MAE \\ \hline
\XSolidBrush & \XSolidBrush & 49.5 & 35.8 & 0.391 & 0.302 \\
\Checkmark & \XSolidBrush & 52.8 & 39.5 & 0.312 & 0.239 \\
\XSolidBrush & \Checkmark & 62.7 & 41.3 & 0.316 & 0.285 \\ \hline
\Checkmark & \Checkmark & \textbf{69.4} & \textbf{54.9} & \textbf{0.295} & \textbf{0.231} \\
\hline
\hline
\end{tabular}
\vspace{-0.5cm}
\label{table4}
\end{table}

\subsubsection{Ablation Study of Different Data Generation Method}
To investigate the effectiveness of each part in our SEKE data generation method, we provide an ablation study for further analysis. Table \ref{table4} presents the test results of models fine-tuned on instruction data generated by different methods. "Prior" denotes the human prior knowledge integrated. When excluded, it indicates that during the generation of missing descriptions, GPT is not provided with any other emotion description knowledge that has already been manually annotated. The absence of SV-UAMC indicates that this strategy is not applied for further reliability enhancement, fully relying on GPT's initial predictions. 

We can observe that both the inclusion of prior knowledge guidance and the SV-UAMC strategy significantly improve the performance of fine-tuned VLLM. 
The incorporation of prior knowledge significantly aids the analysis of continuous emotional dimensions, i.e., valence-arousal. It reveals that accurate annotation of these finer-grained descriptions relies heavily on guidance from other description knowledge and their correlations. 
Further, SV-UAMC contributes to generating more reliable results. Allowing the model to distinguish between confirmed and uncertain parts significantly enhances the quality of instruction data, particularly in expression recognition and AU detection. 
The integration of both components achieves the best performance. This validates the efficacy of our SEKE method, which employs the dual reliability enhancement measures of human prior knowledge guidance and SV-UAMC, generating reliable and comprehensive emotional annotations.

\subsection{Going Deeper into SEKE} 
\subsubsection{The Significance of Complete Description Labels}
\begin{figure}[tb]
\begin{center} 
\includegraphics[width=3in]{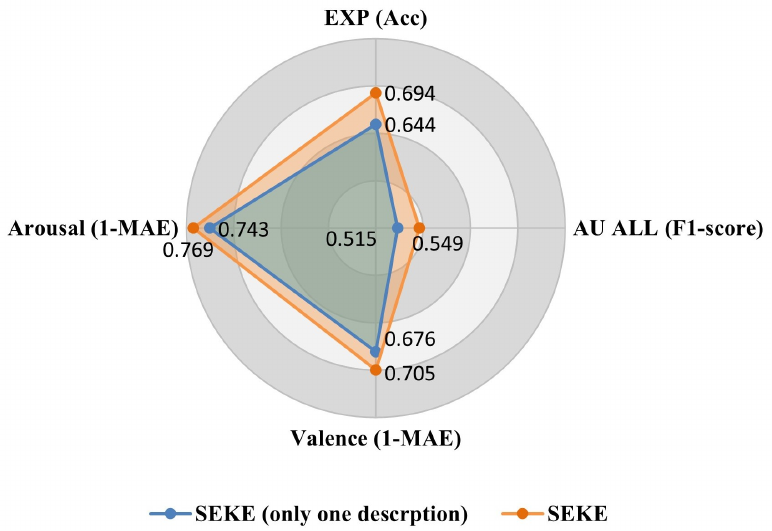}
\end{center}
\caption{Comparison of performance when emotion descriptions are whether complete. Expression accuracy, AU average F1 score, and 1-MAE for valence/arousal on FEAB. SEKE (only one description) denotes the model fine-tuned with only the description label corresponding to the current task.}
\label{eval_discription}
\vspace{-0.3cm}
\end{figure}

To investigate the necessity of maintaining complete three-dimensional emotional descriptions, we conduct controlled experiments. Specifically, three subsets are extracted from the FEID dataset, each retaining annotations for only one type of emotion description. Notably, the retained annotations are fully consistent with the original. Then, we train separate models using each subset to evaluate their performance on corresponding tasks. As shown in Figure \ref{eval_discription}, the comparison reveals the impact of annotation completeness on model efficacy, where a larger radar plot area corresponds to higher accuracy. The `SEKE (only one description)' represents models trained exclusively on single descriptions relevant to their target task.

The results reveal a consistent performance decline across all three tasks when models lack access to complementary descriptions and their correlation knowledge. This demonstrates the necessity of integrating different granularity levels of emotion information. Their interconnected correlations constitute critical affective knowledge, and the learning of it holistically strengthens emotional comprehension. 
Our FEID, the first visual emotion instruction dataset where every sample includes multi-grained emotion descriptions, systematically incorporates such correlation knowledge and enables more precise emotion analysis.

\begin{figure}[tb]
\begin{center} 
\includegraphics[width=3in]{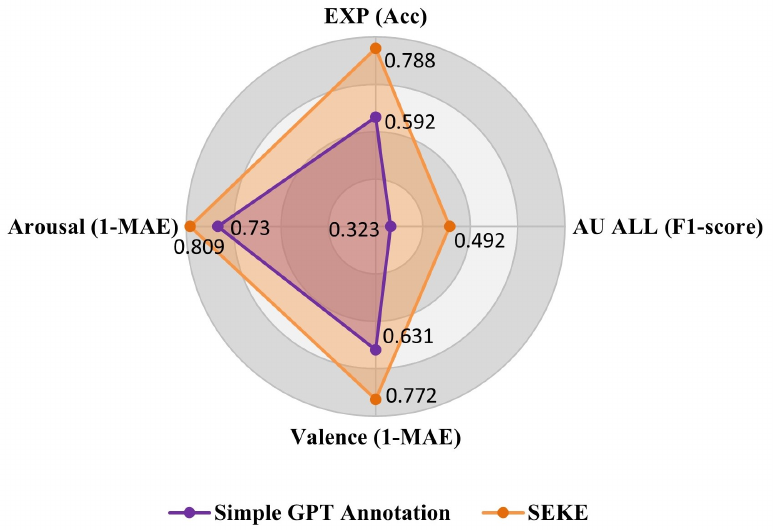}
\end{center}
\caption{Comparison of the reliability of annotated missing labels. Expression accuracy, AU average F1 score, and 1-MAE for valence/arousal on Aff-wild2.}
\label{eval_affwild2}
\vspace{-0.3cm}
\end{figure}

\begin{figure*}[t]
    \centering
    \begin{subfigure}{0.49\textwidth}
        \centering
        \includegraphics[width=\textwidth]{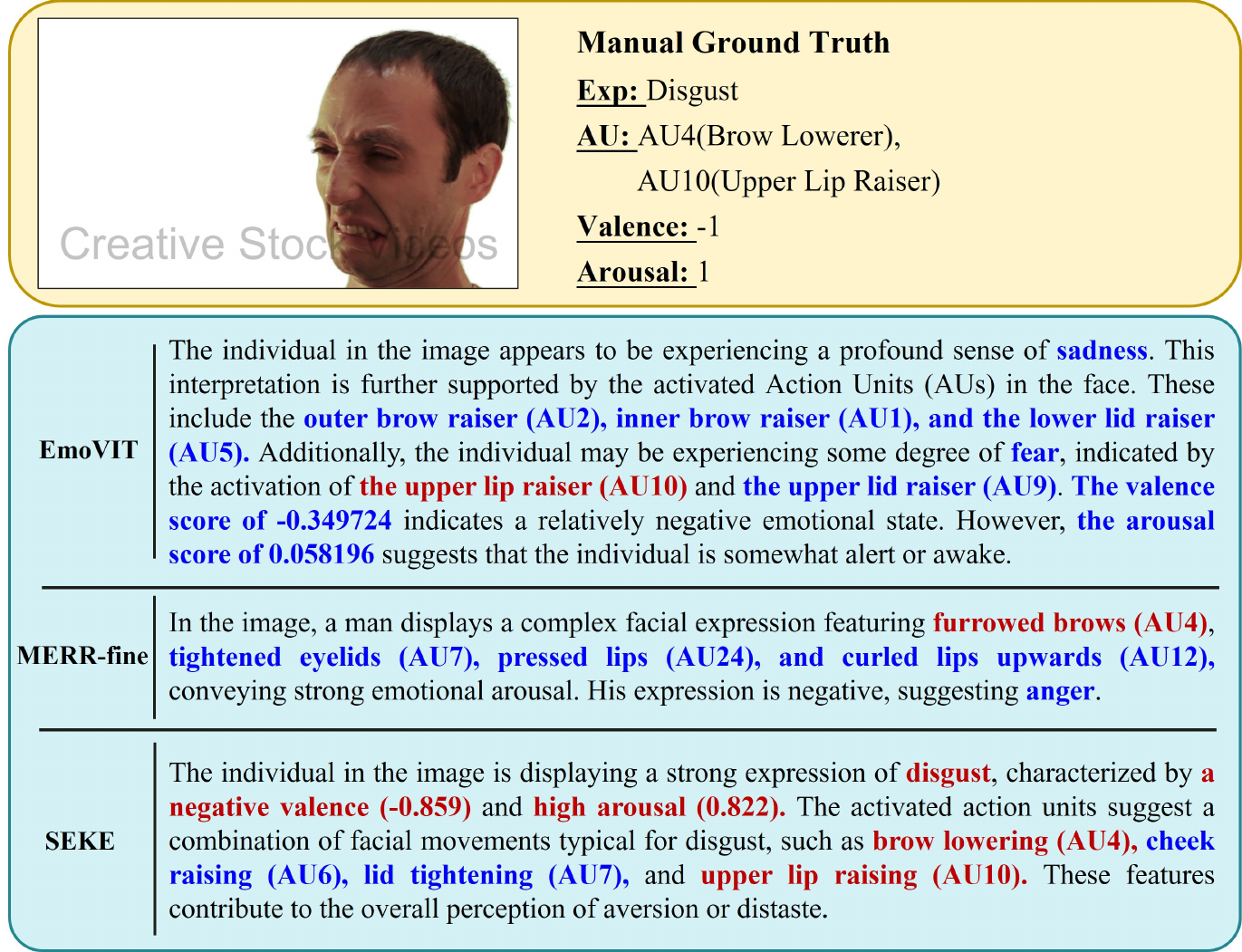}
        \caption{}
    \end{subfigure}
    \begin{subfigure}{0.49\textwidth}
        \centering
        \includegraphics[width=\textwidth]{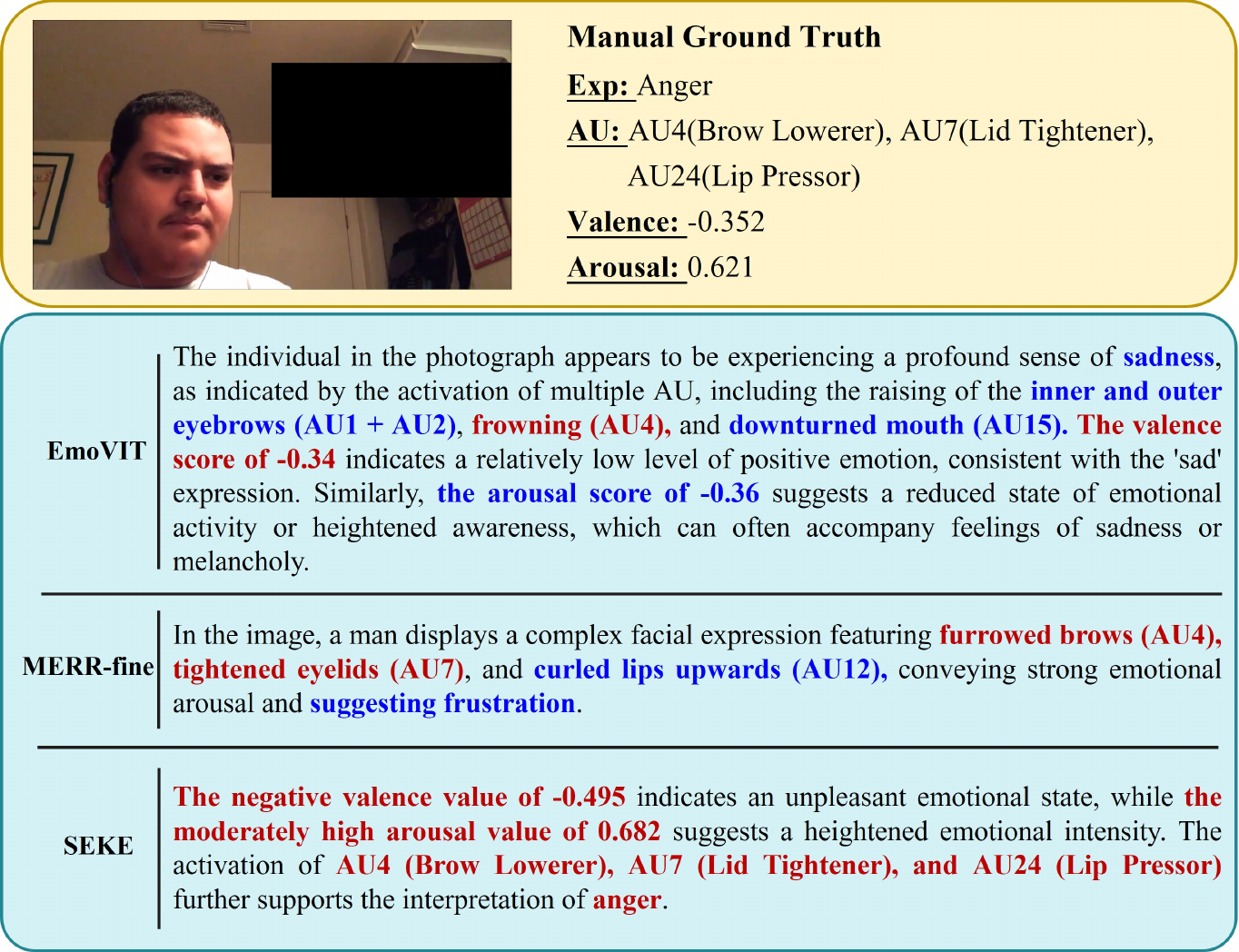}
        \caption{}
    \end{subfigure}
\caption{Two examples comparing the comprehensive emotional reasoning of the SEKE model with that of models fine-tuned on other instruction datasets. Incorrect inferences are marked in blue, and correct in red. For valence and arousal estimates, a prediction is considered correct if the error from the ground truth is within 0.2.}
\label{visual}
\vspace{-0.3cm}
\end{figure*}

\subsubsection{The Reliability of SEKE-Generated Missing Descriptions}
To more intuitively validate the reliability of missing emotion descriptions completed by our SEKE method, we conduct experiments using the Aff-wild2 dataset. Specifically, we select samples containing manual annotations for all three emotion descriptions. For each sample, we mask one description's labels while using the remaining two as prior knowledge to generate the masked one through SEKE, respectively. 
Figure \ref{eval_affwild2} illustrates the alignment between generated descriptions and original manual labels. The `Simple GPT Annotation' refers to directly querying GPT-4o for missing descriptions, representing conventional labeling methods using large models.

The results demonstrate that our SEKE method achieves high accuracy in generating missing emotion descriptions, with MAE for fine-grained valence and arousal estimation remaining below 0.25. Furthermore, SEKE significantly outperforms conventional large-model data generation approaches, particularly with a 16.9$\%$ improvement for AUs. These findings further validate the reliability of our generated instruction data through SEKE, which are in alignment with human expert knowledge.


\subsection{Visualization Analysis}
To more intuitively assess the efficacy of our FEID instruction data, we present two visualization cases comparing the comprehensive emotional reasoning of the SEKE model against models fine-tuned
on other instruction datasets. All experiments utilize LLaVA-OneVision as the base model to ensure consistency. Figure \ref{visual} displays the results. 

The two cases display disgust and anger, emotions frequently conflated with other negative states. Accurate analysis for these samples necessitates integrating both coarse- and fine-grained emotion descriptions to enable robust joint reasoning.
Although models fine-tuned on EmoVIT and MERR-fine can generate complete analyses, their insufficient focus on finer-grained emotion descriptions undermines the accurate analysis of such information, which consequently compromises the basic emotions classification.
After fine-tuning on FEID, our SEKE model effectively learns to use joint reasoning across different emotional dimensions and maintains their correct correlations, thus achieving accurate analysis. These examples demonstrate the efficacy of our instruction data in enabling VLLMs to excel in analyzing both coarse- and fine-grained emotional clues. Beyond visualizing model reasoning, Appendix D provides additional visualization examples of generated instruction data, illustrating variations across different generation methods.
\section{Conclusion}
In this paper, we introduce a self-verification approach to embed human prior emotion knowledge and generate high-quality instruction data for facial emotion analysis. 
The embedding of human prior knowledge guides the large model's automatic annotation rather than relying solely on its raw knowledge.
The proposed SV-UAMC further makes it possible to generate more reliable annotations in a low-cost manner. 
Building on this, we release a high-quality facial emotion instruction dataset, which features multi-grained emotional descriptions and the knowledge of their correlations, effectively enhancing the comprehensive facial emotion analysis capabilities of VLLMs. Besides, we establish a benchmark to evaluate VLLM's effectiveness in this domain. 
To our knowledge, this is the first open-source, coarse- and fine-grained instruction dataset for facial emotion analysis that encompasses expression recognition, valence-arousal estimation, and AU detection simultaneously. 
This work not only advances the capabilities of emotion recognition systems but also paves the way for future research in multimodal emotion analysis. 
\section*{Ethical impact}
 The approach in this paper can benefit many applications, including human-computer interactions. The datasets in this paper are made publicly available with a license that allows free usage for research. Different from facial recognition, the risk of facial emotion analysis is minimal.

\bibliographystyle{ACM-Reference-Format}
\bibliography{sample-base}

\begin{thebibliography}{10}

\bibitem{adyapady2023comprehensive}
R~Rashmi Adyapady and B~Annappa.
\newblock A comprehensive review of facial expression recognition techniques.
\newblock {\em Multimedia Systems}, 29(1):73--103, 2023.

\bibitem{llama3modelcard}
AI@Meta.
\newblock Llama 3 model card.
\newblock 2024.

\bibitem{Anthropic2024Claude}
Anthropic.
\newblock Claude-3.5.
\newblock \url{https://www.anthropic.com/news/claude-3-5-sonnet}, 2024.

\bibitem{bai2025qwen2}
Shuai Bai, Keqin Chen, Xuejing Liu, Jialin Wang, Wenbin Ge, Sibo Song, Kai Dang, Peng Wang, Shijie Wang, Jun Tang, et~al.
\newblock Qwen2. 5-vl technical report.
\newblock {\em arXiv preprint arXiv:2502.13923}, 2025.

\bibitem{ben2021video}
Xianye Ben, Yi~Ren, Junping Zhang, Su-Jing Wang, Kidiyo Kpalma, Weixiao Meng, and Yong-Jin Liu.
\newblock Video-based facial micro-expression analysis: A survey of datasets, features and algorithms.
\newblock {\em IEEE transactions on pattern analysis and machine intelligence}, 44(9):5826--5846, 2021.

\bibitem{Chang_2022_CVPR}
Yanan Chang and Shangfei Wang.
\newblock Knowledge-driven self-supervised representation learning for facial action unit recognition.
\newblock In {\em Proceedings of the IEEE/CVF Conference on Computer Vision and Pattern Recognition (CVPR)}, pages 20417--20426, June 2022.

\bibitem{chen2023minigpt}
Jun Chen, Deyao Zhu, Xiaoqian Shen, Xiang Li, Zechun Liu, Pengchuan Zhang, Raghuraman Krishnamoorthi, Vikas Chandra, Yunyang Xiong, and Mohamed Elhoseiny.
\newblock Minigpt-v2: large language model as a unified interface for vision-language multi-task learning.
\newblock {\em arXiv preprint arXiv:2310.09478}, 2023.

\bibitem{chen2024far}
Zhe Chen, Weiyun Wang, Hao Tian, Shenglong Ye, Zhangwei Gao, Erfei Cui, Wenwen Tong, Kongzhi Hu, Jiapeng Luo, Zheng Ma, et~al.
\newblock How far are we to gpt-4v? closing the gap to commercial multimodal models with open-source suites.
\newblock {\em arXiv preprint arXiv:2404.16821}, 2024.

\bibitem{cheng2024emotion}
Zebang Cheng, Zhi-Qi Cheng, Jun-Yan He, Jingdong Sun, Kai Wang, Yuxiang Lin, Zheng Lian, Xiaojiang Peng, and Alexander Hauptmann.
\newblock Emotion-llama: Multimodal emotion recognition and reasoning with instruction tuning.
\newblock {\em arXiv preprint arXiv:2406.11161}, 2024.

\bibitem{NEURIPS2020_a51fb975}
Zijun Cui, Tengfei Song, Yuru Wang, and Qiang Ji.
\newblock Knowledge augmented deep neural networks for joint facial expression and action unit recognition.
\newblock In H.~Larochelle, M.~Ranzato, R.~Hadsell, M.F. Balcan, and H.~Lin, editors, {\em Advances in Neural Information Processing Systems}, volume~33, pages 14338--14349. Curran Associates, Inc., 2020.

\bibitem{cui2020knowledge}
Zijun Cui, Tengfei Song, Yuru Wang, and Qiang Ji.
\newblock Knowledge augmented deep neural networks for joint facial expression and action unit recognition.
\newblock {\em Advances in Neural Information Processing Systems}, 33:14338--14349, 2020.

\bibitem{depeweg2018decomposition}
Stefan Depeweg, Jose-Miguel Hernandez-Lobato, Finale Doshi-Velez, and Steffen Udluft.
\newblock Decomposition of uncertainty in bayesian deep learning for efficient and risk-sensitive learning.
\newblock In {\em International conference on machine learning}, pages 1184--1193. PMLR, 2018.

\bibitem{ekman1978manual}
Paul Ekman and Wallace~V Friesen.
\newblock {\em Manual for the facial action coding system}.
\newblock Consulting Psychologists Press, 1978.

\bibitem{ekman1987universals}
Paul Ekman, Wallace~V Friesen, Maureen O'sullivan, Anthony Chan, Irene Diacoyanni-Tarlatzis, Karl Heider, Rainer Krause, William~Ayhan LeCompte, Tom Pitcairn, Pio~E Ricci-Bitti, et~al.
\newblock Universals and cultural differences in the judgments of facial expressions of emotion.
\newblock {\em Journal of personality and social psychology}, 53(4):712, 1987.

\bibitem{ekman1997face}
Rosenberg Ekman.
\newblock {\em What the face reveals: Basic and applied studies of spontaneous expression using the Facial Action Coding System (FACS)}.
\newblock Oxford University Press, USA, 1997.

\bibitem{fan2023selfme}
Xinqi Fan, Xueli Chen, Mingjie Jiang, Ali~Raza Shahid, and Hong Yan.
\newblock Selfme: Self-supervised motion learning for micro-expression recognition.
\newblock In {\em Proceedings of the IEEE/CVF Conference on Computer Vision and Pattern Recognition}, pages 13834--13843, 2023.

\bibitem{guo2024stimuvar}
Yuxiang Guo, Faizan Siddiqui, Yang Zhao, Rama Chellappa, and Shao-Yuan Lo.
\newblock Stimuvar: Spatiotemporal stimuli-aware video affective reasoning with multimodal large language models.
\newblock {\em arXiv preprint arXiv:2409.00304}, 2024.

\bibitem{huang2024opera}
Qidong Huang, Xiaoyi Dong, Pan Zhang, Bin Wang, Conghui He, Jiaqi Wang, Dahua Lin, Weiming Zhang, and Nenghai Yu.
\newblock Opera: Alleviating hallucination in multi-modal large language models via over-trust penalty and retrospection-allocation.
\newblock In {\em Proceedings of the IEEE/CVF Conference on Computer Vision and Pattern Recognition}, pages 13418--13427, 2024.

\bibitem{jiang2022disentangling}
Jing Jiang and Weihong Deng.
\newblock Disentangling identity and pose for facial expression recognition.
\newblock {\em IEEE Transactions on Affective Computing}, 13(4):1868--1878, 2022.

\bibitem{kaplan2020scaling}
Jared Kaplan, Sam McCandlish, Tom Henighan, Tom~B Brown, Benjamin Chess, Rewon Child, Scott Gray, Alec Radford, Jeffrey Wu, and Dario Amodei.
\newblock Scaling laws for neural language models.
\newblock {\em arXiv preprint arXiv:2001.08361}, 2020.

\bibitem{kendall2017uncertainties}
Alex Kendall and Yarin Gal.
\newblock What uncertainties do we need in bayesian deep learning for computer vision?
\newblock {\em Advances in neural information processing systems}, 30, 2017.

\bibitem{kollias2019expression}
Dimitrios Kollias and Stefanos Zafeiriou.
\newblock Expression, affect, action unit recognition: Aff-wild2, multi-task learning and arcface.
\newblock {\em arXiv preprint arXiv:1910.04855}, 2019.

\bibitem{li2024llava2}
Bo~Li, Yuanhan Zhang, Dong Guo, Renrui Zhang, Feng Li, Hao Zhang, Kaichen Zhang, Yanwei Li, Ziwei Liu, and Chunyuan Li.
\newblock Llava-onevision: Easy visual task transfer.
\newblock {\em arXiv preprint arXiv:2408.03326}, 2024.

\bibitem{li2024llava}
Feng Li, Renrui Zhang, Hao Zhang, Yuanhan Zhang, Bo~Li, Wei Li, Zejun Ma, and Chunyuan Li.
\newblock Llava-next-interleave: Tackling multi-image, video, and 3d in large multimodal models.
\newblock {\em arXiv preprint arXiv:2407.07895}, 2024.

\bibitem{li2024llava33}
Feng Li, Renrui Zhang, Hao Zhang, Yuanhan Zhang, Bo~Li, Wei Li, Zejun Ma, and Chunyuan Li.
\newblock Llava-next-interleave: Tackling multi-image, video, and 3d in large multimodal models.
\newblock {\em arXiv preprint arXiv:2407.07895}, 2024.

\bibitem{li2020deep}
Shan Li and Weihong Deng.
\newblock Deep facial expression recognition: A survey.
\newblock {\em IEEE transactions on affective computing}, 13(3):1195--1215, 2020.

\bibitem{li2017reliable}
Shan Li, Weihong Deng, and JunPing Du.
\newblock Reliable crowdsourcing and deep locality-preserving learning for expression recognition in the wild.
\newblock In {\em Proceedings of the IEEE conference on computer vision and pattern recognition}, pages 2852--2861, 2017.

\bibitem{lian2024affectgpt}
Zheng Lian, Haiyang Sun, Licai Sun, Jiangyan Yi, Bin Liu, and Jianhua Tao.
\newblock Affectgpt: Dataset and framework for explainable multimodal emotion recognition.
\newblock {\em arXiv preprint arXiv:2407.07653}, 2024.

\bibitem{lian2024gpt}
Zheng Lian, Licai Sun, Haiyang Sun, Kang Chen, Zhuofan Wen, Hao Gu, Bin Liu, and Jianhua Tao.
\newblock Gpt-4v with emotion: A zero-shot benchmark for generalized emotion recognition.
\newblock {\em Information Fusion}, 108:102367, 2024.

\bibitem{lin2023video}
Bin Lin, Bin Zhu, Yang Ye, Munan Ning, Peng Jin, and Li~Yuan.
\newblock Video-llava: Learning united visual representation by alignment before projection.
\newblock {\em arXiv preprint arXiv:2311.10122}, 2023.

\bibitem{lindsley1951emotion}
Donald~B Lindsley.
\newblock Emotion.
\newblock 1951.

\bibitem{liu2024improved}
Haotian Liu, Chunyuan Li, Yuheng Li, and Yong~Jae Lee.
\newblock Improved baselines with visual instruction tuning.
\newblock In {\em Proceedings of the IEEE/CVF Conference on Computer Vision and Pattern Recognition}, pages 26296--26306, 2024.

\bibitem{liu2024visual}
Haotian Liu, Chunyuan Li, Qingyang Wu, and Yong~Jae Lee.
\newblock Visual instruction tuning.
\newblock {\em Advances in neural information processing systems}, 36, 2024.

\bibitem{liu2022facial}
Meng Liu, Yaocong Duan, Robin~AA Ince, Chaona Chen, Oliver~GB Garrod, Philippe~G Schyns, and Rachael~E Jack.
\newblock Facial expressions elicit multiplexed perceptions of emotion categories and dimensions.
\newblock {\em Current Biology}, 32(1):200--209, 2022.

\bibitem{liu2022mafw}
Yuanyuan Liu, Wei Dai, Chuanxu Feng, Wenbin Wang, Guanghao Yin, Jiabei Zeng, and Shiguang Shan.
\newblock Mafw: A large-scale, multi-modal, compound affective database for dynamic facial expression recognition in the wild.
\newblock In {\em Proceedings of the 30th ACM international conference on multimedia}, pages 24--32, 2022.

\bibitem{lucey2010extended}
Patrick Lucey, Jeffrey~F Cohn, Takeo Kanade, Jason Saragih, Zara Ambadar, and Iain Matthews.
\newblock The extended cohn-kanade dataset (ck+): A complete dataset for action unit and emotion-specified expression.
\newblock In {\em 2010 ieee computer society conference on computer vision and pattern recognition-workshops}, pages 94--101. IEEE, 2010.

\bibitem{mavadati2013disfa}
S~Mohammad Mavadati, Mohammad~H Mahoor, Kevin Bartlett, Philip Trinh, and Jeffrey~F Cohn.
\newblock Disfa: A spontaneous facial action intensity database.
\newblock {\em IEEE Transactions on Affective Computing}, 4(2):151--160, 2013.

\bibitem{mollahosseini2017affectnet}
Ali Mollahosseini, Behzad Hasani, and Mohammad~H Mahoor.
\newblock Affectnet: A database for facial expression, valence, and arousal computing in the wild.
\newblock {\em IEEE Transactions on Affective Computing}, 10(1):18--31, 2017.

\bibitem{niu2019multi}
Xuesong Niu, Hu~Han, Shiguang Shan, and Xilin Chen.
\newblock Multi-label co-regularization for semi-supervised facial action unit recognition.
\newblock In {\em Advances in Neural Information Processing Systems}, pages 909--919, 2019.

\bibitem{openai2024gpt4V}
OpenAI.
\newblock Gpt-4v.
\newblock \url{https://openai.com/index/gpt-4v-system-card/}, 2023.

\bibitem{openai2024gpt4o}
OpenAI.
\newblock Hello gpt-4o.
\newblock \url{https://openai.com/index/hello-gpt-4o}, 2024.

\bibitem{paskaleva2024unified}
Reni Paskaleva, Mykyta Holubakha, Andela Ilic, Saman Motamed, Luc Van~Gool, and Danda Paudel.
\newblock A unified and interpretable emotion representation and expression generation.
\newblock In {\em Proceedings of the IEEE/CVF Conference on Computer Vision and Pattern Recognition}, pages 2447--2456, 2024.

\bibitem{russell1980circumplex}
James~A Russell.
\newblock A circumplex model of affect.
\newblock {\em Journal of personality and social psychology}, 39(6):1161, 1980.

\bibitem{Song_Chen_Zheng_Ji_2021}
Tengfei Song, Lisha Chen, Wenming Zheng, and Qiang Ji.
\newblock Uncertain graph neural networks for facial action unit detection.
\newblock {\em Proceedings of the AAAI Conference on Artificial Intelligence}, 35(7):5993--6001, 2021.

\bibitem{Song_2021_CVPR}
Tengfei Song, Zijun Cui, Wenming Zheng, and Qiang Ji.
\newblock Hybrid message passing with performance-driven structures for facial action unit detection.
\newblock In {\em Proceedings of the IEEE/CVF Conference on Computer Vision and Pattern Recognition (CVPR)}, pages 6267--6276, June 2021.

\bibitem{team2023gemini}
Gemini Team, Rohan Anil, Sebastian Borgeaud, Yonghui Wu, Jean-Baptiste Alayrac, Jiahui Yu, Radu Soricut, Johan Schalkwyk, Andrew~M Dai, Anja Hauth, et~al.
\newblock Gemini: a family of highly capable multimodal models.
\newblock {\em arXiv preprint arXiv:2312.11805}, 2023.

\bibitem{team2024gemini}
Gemini Team, Petko Georgiev, Ving~Ian Lei, Ryan Burnell, Libin Bai, Anmol Gulati, Garrett Tanzer, Damien Vincent, Zhufeng Pan, Shibo Wang, et~al.
\newblock Gemini 1.5: Unlocking multimodal understanding across millions of tokens of context.
\newblock {\em arXiv preprint arXiv:2403.05530}, 2024.

\bibitem{touvron2023llama}
Hugo Touvron, Thibaut Lavril, Gautier Izacard, Xavier Martinet, Marie-Anne Lachaux, Timoth{\'e}e Lacroix, Baptiste Rozi{\`e}re, Naman Goyal, Eric Hambro, Faisal Azhar, et~al.
\newblock Llama: Open and efficient foundation language models.
\newblock {\em arXiv preprint arXiv:2302.13971}, 2023.

\bibitem{wang2023rethinking}
Hanyang Wang, Bo~Li, Shuang Wu, Siyuan Shen, Feng Liu, Shouhong Ding, and Aimin Zhou.
\newblock Rethinking the learning paradigm for dynamic facial expression recognition.
\newblock In {\em Proceedings of the IEEE/CVF conference on computer vision and pattern recognition}, pages 17958--17968, 2023.

\bibitem{Qwen2VL}
Peng Wang, Shuai Bai, Sinan Tan, Shijie Wang, Zhihao Fan, Jinze Bai, Keqin Chen, Xuejing Liu, Jialin Wang, Wenbin Ge, Yang Fan, Kai Dang, Mengfei Du, Xuancheng Ren, Rui Men, Dayiheng Liu, Chang Zhou, Jingren Zhou, and Junyang Lin.
\newblock Qwen2-vl: Enhancing vision-language model's perception of the world at any resolution.
\newblock {\em arXiv preprint arXiv:2409.12191}, 2024.

\bibitem{wang2024enhancing}
Weiyun Wang, Zhe Chen, Wenhai Wang, Yue Cao, Yangzhou Liu, Zhangwei Gao, Jinguo Zhu, Xizhou Zhu, Lewei Lu, Yu~Qiao, et~al.
\newblock Enhancing the reasoning ability of multimodal large language models via mixed preference optimization.
\newblock {\em arXiv preprint arXiv:2411.10442}, 2024.

\bibitem{xie2024emovit}
Hongxia Xie, Chu-Jun Peng, Yu-Wen Tseng, Hung-Jen Chen, Chan-Feng Hsu, Hong-Han Shuai, and Wen-Huang Cheng.
\newblock Emovit: Revolutionizing emotion insights with visual instruction tuning.
\newblock In {\em Proceedings of the IEEE/CVF Conference on Computer Vision and Pattern Recognition}, pages 26596--26605, 2024.

\bibitem{yang2024qwen2}
An~Yang, Baosong Yang, Binyuan Hui, Bo~Zheng, Bowen Yu, Chang Zhou, Chengpeng Li, Chengyuan Li, Dayiheng Liu, Fei Huang, et~al.
\newblock Qwen2 technical report.
\newblock {\em arXiv preprint arXiv:2407.10671}, 2024.

\bibitem{Yang_2024_CVPR}
Dingkang Yang, Kun Yang, Mingcheng Li, Shunli Wang, Shuaibing Wang, and Lihua Zhang.
\newblock Robust emotion recognition in context debiasing.
\newblock In {\em Proceedings of the IEEE/CVF Conference on Computer Vision and Pattern Recognition (CVPR)}, pages 12447--12457, June 2024.

\bibitem{yao2024minicpm}
Yuan Yao, Tianyu Yu, Ao~Zhang, Chongyi Wang, Junbo Cui, Hongji Zhu, Tianchi Cai, Haoyu Li, Weilin Zhao, Zhihui He, et~al.
\newblock Minicpm-v: A gpt-4v level mllm on your phone.
\newblock {\em arXiv preprint arXiv:2408.01800}, 2024.

\bibitem{zhai2023sigmoid}
Xiaohua Zhai, Basil Mustafa, Alexander Kolesnikov, and Lucas Beyer.
\newblock Sigmoid loss for language image pre-training.
\newblock In {\em Proceedings of the IEEE/CVF International Conference on Computer Vision}, pages 11975--11986, 2023.

\bibitem{zhang2013high}
Xing Zhang, Lijun Yin, Jeffrey~F Cohn, Shaun Canavan, Michael Reale, Andy Horowitz, and Peng Liu.
\newblock A high-resolution spontaneous 3d dynamic facial expression database.
\newblock In {\em 2013 10th IEEE international conference and workshops on automatic face and gesture recognition (FG)}, pages 1--6. IEEE, 2013.

\bibitem{zhang2014bp4d}
Xing Zhang, Lijun Yin, Jeffrey~F Cohn, Shaun Canavan, Michael Reale, Andy Horowitz, Peng Liu, and Jeffrey~M Girard.
\newblock Bp4d-spontaneous: a high-resolution spontaneous 3d dynamic facial expression database.
\newblock {\em Image and Vision Computing}, 32(10):692--706, 2014.

\bibitem{zhao2024khfa}
Huijuan Zhao, Shuangjiang He, Congju Du, Linyun Liu, and Li~Yu.
\newblock Khfa: Knowledge-driven hierarchical feature alignment framework for subject-invariant facial action unit detection.
\newblock {\em IEEE Transactions on Instrumentation and Measurement}, 2024.

\end{thebibliography}

\end{document}